\documentclass[runningheads]{llncs}

\usepackage{eccv}

\usepackage{eccvabbrv}

\usepackage{graphicx}
\usepackage{booktabs}
\usepackage{algorithm,algorithmicx,algpseudocode}
\usepackage{multirow}
\usepackage{makecell}

\usepackage[accsupp]{axessibility}  

\usepackage{hyperref}

\usepackage{orcidlink}

\begin{document}

\title{AdvDiff: Generating Unrestricted Adversarial Examples using Diffusion Models} 

\titlerunning{AdvDiff}

\author{Xuelong Dai\inst{1}\orcidlink{0000-0001-6646-6514} \and
Kaisheng Liang\inst{1}\orcidlink{0000-0002-8297-6378} \and
Bin Xiao\inst{1}\orcidlink{0000-0003-4223-8220}}

\authorrunning{X.~Dai et al.}

\institute{
The Hong Kong Polytechnic University\\
\email{xuelong.dai@connect.polyu.hk, cskliang@comp.polyu.edu.hk, b.xiao@polyu.edu.hk }}

\maketitle

\begin{abstract}
Unrestricted adversarial attacks present a serious threat to deep learning models and adversarial defense techniques. They pose severe security problems for deep learning applications because they can effectively bypass defense mechanisms. However, previous attack methods often directly inject Projected Gradient Descent (PGD) gradients into the sampling of generative models, which are not theoretically provable and thus generate unrealistic examples by incorporating adversarial objectives, especially for GAN-based methods on large-scale datasets like ImageNet. In this paper, we propose a new method, called AdvDiff, to generate unrestricted adversarial examples with diffusion models. We design two novel adversarial guidance techniques to conduct adversarial sampling in the reverse generation process of diffusion models. These two techniques are effective and stable in generating high-quality, realistic adversarial examples by integrating gradients of the target classifier interpretably. Experimental results on MNIST and ImageNet datasets demonstrate that AdvDiff is effective in generating unrestricted adversarial examples, which outperforms state-of-the-art unrestricted adversarial attack methods in terms of attack performance and generation quality.

\end{abstract}

\section{Introduction}

While the deep learning (DL) community continues to explore the wide range of applications of DL models, researchers \cite{szegedy2013intriguing} have demonstrated that these models are highly susceptible to deception by adversarial examples. Adversarial examples are generated by adding perturbations to clean data. The perturbed examples can deceive DL classifiers with high confidence while remaining imperceptible to humans. Many strong attack methods \cite{madry2017towards,dong2018boosting,carlini2017towards,croce2020reliable,liang2023styless,li2023physical} are proposed and investigated to improve the robustness of DL models.

In contrast to existing perturbation-based adversarial attacks, Song et al. \cite{song2018constructing}
found that using a well-trained generative adversarial network with an auxiliary classifier (AC-GAN) \cite{odena2017conditional} can directly generate new adversarial examples without perturbing the clean data. These newly generated examples are considered \textbf{unrestricted} as they are obtained by optimizing input noise vectors without any norm restrictions. Compared to traditional adversarial examples, unrestricted adversarial examples \cite{deb2020advfaces,qiu2020semanticadv} are more aggressive against current adversarial defenses. A malicious adversary can also generate an unlimited number of unrestricted adversarial examples using a trained GAN.

Diffusion models \cite{ho2020denoising} are likelihood-based generative models proposed recently, which emerged as a strong competitor to GANs. Diffusion models have outperformed GANs for image synthesis tasks \cite{dhariwal2021diffusion,rombach2022high,kim2022diffusionclip}. Compared with GAN models, diffusion models are more stable during training and provide better distribution coverage. Diffusion models contain two processes: a forward diffusion process and a reverse generation process. The forward diffusion process gradually adds Gaussian noise to the data and eventually transforms it into noise. The reverse generation process aims to recover the data from the noise by a denoising-like technique. A well-trained diffusion model is capable of generating images with random noise input. Similar to GAN models, diffusion models can achieve adversarial attacks by incorporating adversarial objectives \cite{chen2023advdiffuser,chen2023content,chen2023diffusion}.

 GAN-based unrestricted adversarial attacks often exhibit poor 
 performance on high-quality datasets, particularly in terms of visual quality, because they directly add the PGD perturbations to the GAN latents without theoretic supports. These attacks tend to generate low-quality adversarial examples compared to benign GAN examples \cite{song2018constructing}. Therefore, these attacks are not imperceptible among GAN synthetic data. 
 Diffusion models, however, offer state-of-the-art generation performance \cite{dhariwal2021diffusion} on challenging datasets like LSUN \cite{yu2015lsun} and ImageNet \cite{deng2009imagenet}. The conditional diffusion models can generate images based on specific conditions by sampling from a perturbed conditional Gaussian noise, which can be carefully modified with adversarial objectives. These properties make diffusion models more suitable for conducting unrestricted adversarial attacks. Nevertheless, existing adversarial attack methods using diffusion models \cite{chen2023advdiffuser,chen2023content,chen2023diffusion} adopt similar PGD perturbations to the sample in each reverse generation process, making them generate relatively low-quality adversarial examples.  

\begin{figure}[t]
   \begin{center}
     \includegraphics[width=1.0\linewidth]{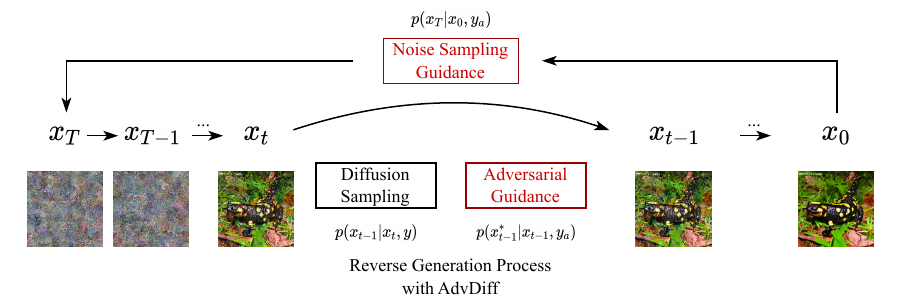}
   \end{center}
      \caption{\textbf{The two new guidance techniques in our AdvDiff to generate unrestricted adversarial examples.} During the reverse generation process, the \emph{adversarial guidance} is added at timestep $x_t$, which injects the adversarial objective $y_a$ into the diffusion process. The \emph{noise sampling guidance} modifies the original noise by increasing the conditional likelihood of $y_a$.}
   \label{fig:pip}
   \end{figure}

In this paper, we propose a novel and interpretable unrestricted adversarial attack method called AdvDiff that utilizes diffusion models for adversarial examples generation, as shown in Figure \ref{fig:pip}. Specifically, AdvDiff uses a trained conditional diffusion model to conduct adversarial attacks with two new adversarial guidance techniques. 1) During the reverse generation process, we gradually add \emph{adversarial guidance} by increasing the likelihood of the target attack label. 2) We perform the reverse generation process multiple times, adding adversarial prior knowledge to the initial noise with the \emph{noise sampling guidance}.

Our theoretical analysis indicates that these adversarial guidance techniques can effectively craft adversarial examples by the reverse generation process with adversarial conditional sampling. Furthermore, the sampling of AdvDiff benefits from stable and high sample quality of the diffusion models sampling, which leads to the generation of realistic unrestricted adversarial examples.
Through extensive experiments conducted on two datasets, i.e., the high-quality dataset ImageNet, and the small, robust dataset MNIST, we have observed a significant improvement in the attack performance using  AdvDiff with diffusion models. 
These results prove that our proposed AdvDiff is more effective than previous unrestricted adversarial attack methods in conducting unrestricted adversarial attacks to generate high-fidelity and diverse examples without decreasing the generation quality.

Our contributions can be summarized as follows:

\begin{itemize}
\setlength{\itemsep}{0pt}

\setlength{\parsep}{0pt}

\setlength{\parskip}{0pt}
   \item We propose AdvDiff, the new form unrestricted adversarial attack method that utilizes the reverse generation process of diffusion models to generate realistic adversarial examples.
   \item We design two new effective adversarial guidance techniques to the sampling process that incorporate adversarial objectives to the diffusion model without re-training the model. Theoretical analysis reveals that AdvDiff can generate unrestricted adversarial examples while preserving the high-quality and stable sampling of the conditional diffusion models.
   \item  We perform extensive experiments to demonstrate that AdvDiff achieves an overwhelmingly better performance than GAN models on unrestricted adversarial example generation.

\end{itemize}

\section{Preliminaries}
In this section, we introduce the diffusion model and the classifier guidance for constructing our adversarial diffusion model.

\subsection{Diffusion Model}
The Denoising Diffusion Probabilistic Model (DDPM) \cite{ho2020denoising} is utilized to learn the Gaussian
transitions from $p(x_t)=\mathcal{N}(x_t;0,\mathbf{I})$ to recover the data $x_0 \sim q(x_0)$ with a Markov chain. We call this denoising-like process the \textit{reverse generation process}. Following the pre-defined $T$ time steps, DDPM obtains a sequence of noisy data $\{x_{T-1}, \dots, x_{1}\}$ and finally recovers the data $x_0$. It is defined as:
\begin{equation}
    p_\theta(x_{t-1}|x_t) := \mathcal{N}(x_{t-1}:\mu_{\theta}(x_t,t),\Sigma_{\theta}(x_t,t)) 
\end{equation}

Conversely, the \textit{forward diffusion process} leverages a fixed Markov chain to iteratively add Gaussian noise to the sampled data $x_0 \sim q(x_0)$ according to the scheduling function $\beta_1, \dots, \beta_N$. Specifically, it is defined as:
\begin{equation}
    q(x_t|x_{t-1}):=\mathcal{N}(x_t:\sqrt[]{1-\beta_t}x_{t-1},\beta_t\textbf{I} ) 
\end{equation}

Training of the DDPM requires accurate inference of the mean value $\mu_{\theta}$ with a deep learning model in the reverse generation process. The objective is to learn the variational lower-bound (VLB) on $\log p_\theta(x_0)$. In order to complete the training objective, we train the model $\epsilon_\theta$ to predict the added Gaussian noise in the forward diffusion process. The standard mean-squared error loss is adopted:
\begin{equation}
    \mathcal{L}_{DDPM}:= E_{t\sim [1,T],\epsilon \sim\mathcal{N}(0,\textbf{I})}\left \| \epsilon-\epsilon_\theta (x_t,t) \right \|^2 
\end{equation}

Song et al. \cite{song2020denoising} proposed DDIM, which provides an alternative noising process without restricting to a Markov chain. DDIM can achieve a much faster sampling than DDPM with the same training as DDPM. We perform experiments on both DDPM and DDIM to demonstrate the usability of our AdvDiff. DDPM is adopted to introduce our method for simplicity.

\subsection{Classifier-Guided Guidance}
Dhariwal et al. \cite{dhariwal2021diffusion} achieved conditional diffusion sampling by adopting a trained classifier. The conditional information is injected into the diffusion model by modifying the mean value $\mu_{\theta}(x_t,t)$ of the samples according to the gradient of the prediction of the target class $y$ by the trained classifier. They adopted log probability to calculate the gradient, and the mean value is given by:
\begin{equation}
    \hat{\mu}_{\theta}(x_t,t) = \mu_{\theta}(x_t,t) + s\cdot \nabla_{x_t}\log p_{\phi}(y|x_t) 
\end{equation}
where $s$ is the guidance scale.

\subsection{Classifier-Free Guidance}
Ho et al. \cite{ho2021classifier} recently proposed a new conditional diffusion model using classifier-free guidance that injects class information without adopting an additional classifier. The classifier-free guidance utilizes a conditional diffusion model $p_{\theta}(x|y)$ for image synthesis with given labels. For effective training, they jointly train the unconditional diffusion model $p_{\theta}(x|\emptyset)$ and the conditional diffusion model $p_{\theta}(x|y)$, where the unconditional diffusion model is simply replacing the label information with $\emptyset$. Sampling is performed by pushing the model towards the latent space of  $p_{\theta}(x|y)$ and away from $p_{\theta}(x|\emptyset)$:
\begin{equation}
    \hat{\epsilon}_{\theta }(x_t|y)= {\epsilon}_{\theta }(x_t|\emptyset) + w\cdot({\epsilon}_{\theta }(x_t|y)-{\epsilon}_{\theta }(x_t|\emptyset))
\end{equation}
where $w$ is the weight parameter for class guidance and $\emptyset$ is the empty set. 

The idea of classifier-free guidance is inspired by the gradient of an implicit classifier $p^i(y|x)\propto p(x|y)/p(x)$, the gradient of the classifier would be:
\begin{align}
\nabla_x logp^i(y|x)&\propto  \nabla_x logp(x|y)-\nabla_x logp(x) \notag \\
&\propto {\epsilon}_{\theta }(x_t|y)-{\epsilon}_{\theta }(x_t|\emptyset)
\end{align}

The classifier-free guidance has a good capability of generating high-quality conditional images, which is critical for performing adversarial attacks. The generation of these images does not rely on a classification model and thus can better fit the conditional distribution of the data.

\section{Adversarial Diffusion Sampling}

\subsection{Rethinking Unrestricted Adversarial Examples}

Song et al. \cite{song2018constructing} presented a new form of adversarial examples called unrestricted adversarial examples (UAEs). These adversarial examples are not generated by adding perturbations over the clean data but are directly generated by any generative model. UAEs can be viewed as false negative errors in the classification tasks, and they can also bring server security problems to deep learning models. These generative-based UAEs can be formulated as:

\begin{equation}
   A_{\text{UAE}} \triangleq \{x \in \mathcal{G}(z_{\text{adv}},y)|y \neq f(x)\}
\end{equation}
where $f(\cdot)$ is the target model for unrestricted adversarial attacks. The unrestricted adversarial attacks aim to generate UAEs that fool the target model while still can be visually perceived as the image from ground truth label $y$.

Previous UAE works adopt GAN models for the generation of UAEs, and these works perturb the GAN latents by maximizing the cross-entropy loss of the target model, i.e., $\max_{z_{\text{adv}}}\mathcal{L}(f(\mathcal{G}(z_{\text{adv}},y)),y)$. Ideally, the generated UAEs should guarantee similar generation quality to the samples crafted by standard $z$ because successful adversarial examples should be imperceptible to humans. In other words, UAEs should not be identified among the samples with adversarial latents and standard latents. 

However, due to GAN's poor interpretability, there's no theoretical support on $z_{adv}$ that can craft UAEs with normally trained GANs. The generator of GAN is not trained with $z_{adv} = z + \nabla\mathcal{L}$ but only $z \sim \mathcal{N}(0, \textbf{I})$. Therefore, GAN-based UAEs encounter a significant decrease in generation quality because samples with $z_{adv}$ are not well-trained compared with samples with $z \sim \mathcal{N}(0, \textbf{I})$. Moreover, the GAN latents are sampled from low dimensional latent spaces. Therefore, GANs are extremely sensitive to the latent $z$ \cite{shen2020interpreting,liang2022exploring}. If we inject gradients of the classification results into GAN latents, GAN-based methods are more likely to generate flipped-label UAEs (images corresponding to the targeted attack label $y_a$ instead of the conditional generation label $y$) and distorted UAEs. However, these generation issues are hard to address only by attack success rate (ASR). In other words, even with a high ASR, some of the successful UAEs with GAN-based methods should be identified as failure cases for poor visual quality. However, such cases can not be reflected by ASR but can be evaluated by generation quality. All these problems may indicate that GAN models are not suitable for generative-based adversarial attacks.

\begin{figure}[t]
   \begin{center}
     \includegraphics[width=0.4\linewidth]{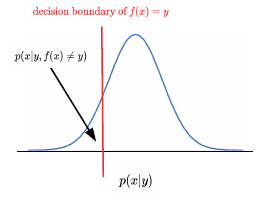}
   \end{center}
      \caption{\textbf{Unrestricted adversarial examples generated by the diffusion model.} The generated adversarial examples should be visually indistinguishable from clean data with label $y$ but wrongly classified by the target classifier $f$.}

   \label{fig:fn}
   \end{figure}

Diffusion models have shown better performance on image generation than GAN models \cite{dhariwal2021diffusion}. They are log-likelihood models with interpretable generation processes. In this paper, we aim to generate UAEs by injecting the adversarial loss with theoretical proof and without sabotaging the benign generation process, where we increase the conditional likelihood on the target attack label by following the diffusion process. The perturbations are gradually injected with the backward generation process of the diffusion model by the same sample procedure. As shown in Figure \ref{fig:fn}, the diffusion model can sample images from the conditional distribution $p(x|y)$. The samples from $p(x|y,f(x)\ne y)$ are the adversarial examples that are misclassified by $f(\cdot)$. These examples also follow the data distribution $p(x|y)$ but on the other side of the label $y$ 's decision boundary of $f(\cdot)$. Moreover, the diffusion model's generation process takes multiple sampling steps. Thus, we don't need one strong perturbation to the latent like GAN-based methods. The AdvDiff perturbations at each step are unnoticeable, and perturbations are added to the high dimensional sampled data rather than low dimensional latents. Therefore, AdvDiff with diffusion models can preserve the generation quality and barely generates flipped-label or distorted UAEs.

\subsection{Adversarial Diffusion Sampling with Theoretical Support}

There are several existing adversarial attack methods \cite{chen2023advdiffuser,chen2023diffusion,chen2023content} that adopt diffusion models to generate adversarial examples. However, these methods still adopt PGD or I-FGSM gradients to perturb the diffusion process for constructing adversarial examples. As discussed earlier, the generation process of diffusion models is a specially designed sampling process from given distributions. Such adversarial gradients change the original generation process and can harm the generation quality of the diffusion model. Additionally, these methods fail to give a comprehensive discussion of the adversarial guidance with theoretical analysis. Therefore, we aim to design a \textbf{general} and \textbf{interpretable} method to generate adversarial examples using diffusion models \textbf{without} affecting the benign diffusion process.

\subsection{Adversarial Guidance}

Inspired by Dhariwal's work \cite{dhariwal2021diffusion} that achieves the conditional image generation by classifier gradient guidance $\nabla_{x_t}\log p_{\phi}(y|x_t)$, we generate our UAEs with adversarial gradient guidance over the reverse generation process. Our attack aims at utilizing a conditional diffusion model $\epsilon_{\theta}(x_t,y)$ to generate $x_0$ fits the ground truth label $y$ while deceiving the target classifier with $p_{f}(x_0) \ne y$. These generated samples are the false negative results in $p_{f}$'s classification results. 

Normally, we will obtain the images with label $y$ by following the standard reverse generation process with classifier-free guidance:
\begin{equation}
{x}_{t-1} = \mu({x}_t,y) + \sigma_t{\varepsilon}
\label{eq:cfg}
\end{equation}
where $\mu({x}_t,y)$ is the conditional mean value and $\varepsilon$ is sampled from ${\varepsilon}\sim \mathcal{N}(0,\textbf{I})$.

Sampling by Equation \ref{eq:cfg}, we obtain the samples with the generation process $p(x_{t-1}|x_t,y)$. Following the above-mentioned definition of UAEs, we can get our adversarial examples by adding adversarial guidance to the standard reverse process, which is performing another sampling with the adversarial generation process $p(x^*_{t-1}|x_{t-1},f(x)\ne y)$. We find that specifying a target label for the adversarial generation process is more effective during experiments. Suggest the target label $y_{a}$ is the target for the adversarial attacks, the adversarial example is sampled by the following steps:
\begin{equation}
{x}^*_{t-1} = x_{t-1} +\sigma_t^2 s\nabla_{{x}_{t-1}} \log p_{f}({y_{a}}|{x}_{t-1})
\label{eq:sample}
\end{equation}
where $s$ is the adversarial guidance scale. The derivation of Equation \ref{eq:sample} is given in Appendix A. Intuitively, the adversarial guidance encourages the generation of samples with a higher likelihood of the target label.

In practice, we utilize the classifier-free guidance to train a conditional diffusion model $\epsilon_{\theta}(\cdot)$ as our basic generation model.

\subsection{Noise Sampling Guidance}

We can improve the reverse process by adding an adversarial label prior to the noise data $x_T$. The UAEs are a subset of the dataset labeled with $y$. They can be viewed as the conditional probability distribution with $p(x|y, f(x)=y_{a})$ during sampling, and $y_a$ is the target label for the adversarial attack. Therefore, we can add the adversarial label prior to $x_t$ with Bayes' theorem:
\begin{align}
    p({x}_{T}|y_{a}) =& \frac{p(y_{a}|{x}_{T})p({x}_{T})}{p(y_{a})} = \frac{p(y_{a}|x_T,x_0)p(x_T|x_0)}{p(y_{a}|x_0)} \notag \\=&p(x_T|x_0)e^{\log p(y_{a}|x_T)- \log p(y_{a}|x_0) }
    \label{eq:noise}
\end{align}

We can infer the $x_t$ with the adversarial prior by Equation \ref{eq:noise}, i.e.,
\begin{equation}
    {x}_{T} = (\mu({x}_0,y) + \sigma_t{\varepsilon}) +\bar{\sigma}_T^2 a\nabla_{{x}_0} \log p_{f}({y_{a}}|{x}_0)
    \label{eq:sample_noise}
\end{equation}
where $a$ is the noise sampling guidance scale. See Appendix B for detailed proof.

Equation \ref{eq:sample_noise} is similar to Equation \ref{eq:sample} as they both add adversarial guidance to the reverse generation process. However, the noise sampling guidance is added to $x_T$ according to the final classification gradient $\nabla_{{x}_0} \log p_{f}({y_{a}}|{x}_0)$, which provides a strong adversarial guidance signal directly to the initial input of the generative model. The gradient of Equation \ref{eq:sample_noise} is effective as it reflects the eventual classification result of the target classifier.

\subsection{Training-Free Adversarial Attack}

The proposed adversarial attack does not require additional modification on the training of the diffusion model. The adversarial examples are sampled by using Algorithm \ref{alg:sampling} over the trained classifier-free diffusion model $\epsilon_{\theta}(\cdot)$.  We give the AdvDiff algorithm on DDIM in the Appendix.

\begin{algorithm}[tb]
  \caption{DDPM Adversarial Diffusion Sampling} 
  \label{alg:sampling}
  \begin{algorithmic}[1]
    \Require $y_a$: target label for adversarial attack
    \Require $y$: ground truth class label
    \Require $s,a$: adversarial guidance scale
    \Require $w$: classification guidance scale
    \Require $N$: noise sampling guidance steps
    \Require $T$: reverse generation process timestep

    \State $x_{T} \sim \mathcal{N}(0, \textbf{I})$
    \State $x_{adv} = \varnothing $
    \For{$i = 1\ldots N$}
    \For{$t=T, \dotsc, 1$}

      \State $\tilde{\epsilon}_t = (1+w)\epsilon_\theta(x_{t}, y) - w\epsilon_{\theta}(x_{t})$ 
      \State Classifier-free sampling $x_{t-1}$ with $\tilde{\epsilon}_t$.
      \State Input $x_{t-1}$ to target model and get the gradient $\log p_{f}({y_{a}}|{x}_{t-1}))$
      \State ${x}^*_{t-1} = x_{t-1} +\sigma_t^2 s\nabla_{{x}_{t-1}} \log p_{f}({y_{a}}|{x}_{t-1})$
    \EndFor
      \State Obtain classification result from $f(x_0)$
      \State Compute the gradient with $\log p_{f}({y_{a}}|{x}_0)$
      \State Update $x_{T}$ by $x_T = x_T + \bar{\sigma}_T^2 a\nabla_{{x}_0} \log p_{f}({y_{a}}|{x}_0)$
      \State  $x_{adv} \gets x_{0}$ if $f(x_0)=y_a$

    \EndFor
    \State \textbf{return} $x_{adv}$
  \end{algorithmic}
\end{algorithm}

\section{Experiments}

\begin{figure}[t]
   \begin{center}
     \includegraphics[width=0.7\linewidth]{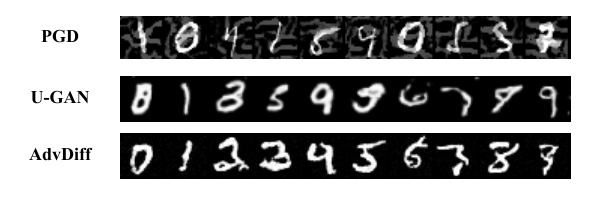}
   \end{center}
      \caption{\textbf{Adversarial examples on the MNIST dataset.} Perturbation-based attack methods generate noise patterns to conduct attacks, while unrestricted adversarial attacks (U-GAN and AdvDiff) are imperceptible to the clean data. }
   \label{fig:mnist}

   \end{figure}

\textbf{Datasets and target models.} We use two datasets for major evaluation: MNIST \cite{deng2012mnist} and ImageNet \cite{deng2009imagenet}. MNIST is a 10-classes dataset consisting of handwritten numbers from 0 to 9. We adopt the MNIST dataset to evaluate our method for low-quality robust image generation. ImageNet is a large visual database with 1000 object classes and is used for the high-quality generation task. For target classifiers, we adopt simple LeNet5 \cite{lecun1998gradient}, and ResNet18 \cite{he2016deep} for the MNIST dataset, and the widely-used ResNet50 \cite{he2016deep} and WideResNet50-2 \cite{zagoruyko2016wide} for the ImageNet dataset.

\textbf{Comparisons.} It is not applicable to give a clear comparison between perturbation attacks and unrestricted attacks because perturbation attacks have the corresponding ground truth while unrestricted attacks do not. We mainly compare our method with the unrestricted adversarial attack U-GAN \cite{song2018constructing} and give the discussion with the AutoAttack \cite{croce2020reliable}, PGD \cite{madry2017towards}, BIM \cite{dong2018boosting}, and C\&W \cite{carlini2017towards} perturbation-based attacks under norm $\ell_\text{inf}=8/255$. For U-GAN, We adopt AC-GAN \cite{odena2017conditional} for the MNIST dataset, and SAGAN \cite{zhang2019self} and BigGAN \cite{brock2018large} for the ImageNet dataset, as AC-GAN has shown poor performance on ImageNet. We use the official code from DiffAttack \cite{chen2023diffusion} and implement AdvDiffuser by ourselves \cite{chen2023advdiffuser} for comparisons. We do not compare with Chen et al. \cite{chen2023content}, because they use a similar method as DiffAttack and without official code. Because existing diffusion model attacks are all untargeted attacks, we include the untargeted version of AdvDiff for a clear comparison, which is represented by ``AdvDiff-Untargeted''. 

\textbf{Implementation details.} Because our adversarial diffusion sampling does not require additional training to the original diffusion model, we use the pre-trained diffusion model in our experiment. We adopt DDPM \cite{ho2020denoising} with classifier-free guidance for the MNIST dataset and Latent Diffusion Model (LDM) \cite{rombach2022high} with DDIM sampler for the ImageNet dataset.  For MNIST dataset, we use $N=10$, $s=0.5$, and $a=1.0$, And $N=5$, $s=0.7$, and $a=0.5$ for ImageNet dataset. More details and experiments are given in the Appendix.

\textbf{Evaluation metrics.} We utilize the top-1 classification result to evaluate the Attack Success Rate (ASR) on different attack methods under untargeted attack settings. As discussed earlier, GAN-based UAEs often encounter severe generation quality drops compared to benign GAN samples. Therefore, we give comparisons of generation performance on ImageNet to evaluate the attack performance of different UAEs in imperceptibly. The results are averaged with five runs. We use ResNet50 as the target model for default settings.

\begin{figure}[t]
   \begin{center}
     \includegraphics[width=1.0\linewidth]{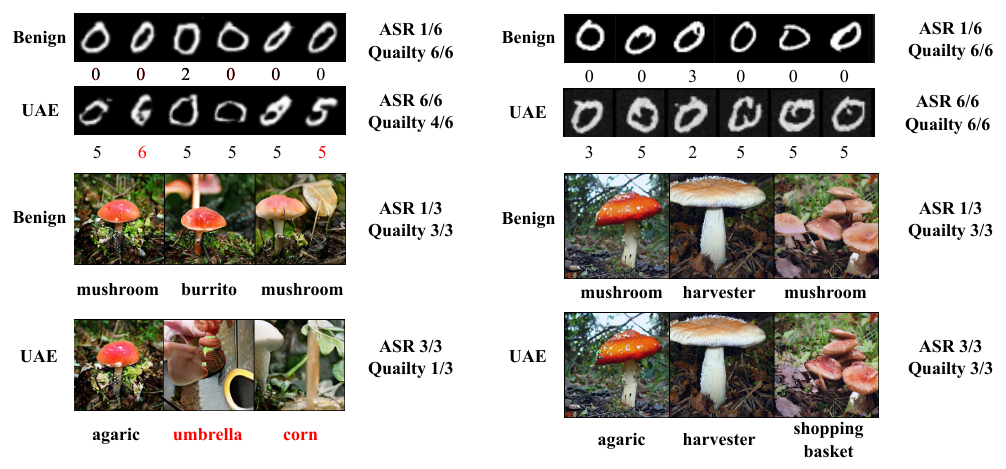}
   \end{center}
      \caption{\textbf{Comparisons of unrestricted adversarial attacks between GANs and diffusion models on two datasets.}. Left: generated samples from U-GAN (BigGAN for ImageNet dataset). Right: generated samples from AdvDiff. We generate unrestricted adversarial examples on the MNIST ``0" label and ImageNet ``mushroom" label. U-GAN is more likely to generate adversarial examples with the target label, i.e., examples with red font. However, AdvDiff tends to generate the ``false negative" samples by the target classifier by combing features from the target label.}
   \label{fig:gvd}

   \end{figure}
   
\subsection{Attack Performance}
\noindent \textbf{MNIST} We show the attack success rate against the normally trained model and adversarially trained model \cite{madry2017towards} in the MNIST dataset. All the selected adversarial attacks achieve over 90\% attack success rate against the normally trained model. The adversarially trained model can effectively defend against perturbation-based adversarial attacks for their noise-like perturbation generation patterns, as reported in Table \ref{tab:mnis}. However, the UAEs obviously perform better with their non-perturbed image generation. Despite the fact that the unrestricted attack can break through the adversarial defenses, the crafted adversarial examples should also be imperceptible to humans for a reasonably successful attack. The visualized adversarial examples in Figure \ref{fig:mnist} show that the perturbation-based adversarial attacks tend to blur the original images while U-GAN can generate mislabeled adversarial examples.

\begin{table}[t]
\parbox{.38\textwidth}{
\centering

\caption{\textbf{The attack success rate on MNIST dataset.}  }
   
\label{tab:mnis}
\resizebox{0.38\textwidth}{!}{  
\begin{tabular}{l|cccc}

\Xhline{3\arrayrulewidth}
\multirow{3}{*}{Method} & \multicolumn{4}{c}{ASR(\%)}                                                                        \\ \cline{2-5} 
                        & \multicolumn{2}{c|}{LeNet5}                                & \multicolumn{2}{c}{ResNet18}          \\
                        & Clean        & \multicolumn{1}{c|}{PGD-AT}                 & Clean        & PGD-AT                 \\ \hline
PGD                     & 99.8         & \multicolumn{1}{c|}{25.6}                   & 99.3         & 20.8                   \\
BIM                     & 99.6         & \multicolumn{1}{c|}{34.6}                   & \textbf{100} & 31.5                   \\
C\&W                    & \textbf{100} & \multicolumn{1}{c|}{68.6}                   & \textbf{100} & 64.5                   \\ \hline
U-GAN                   & 88.5         & \multicolumn{1}{c|}{\textit{79.4}}          & 85.6         & \textit{75.1}          \\
AdvDiff              & 94.2         & \multicolumn{1}{c|}{\textit{\textbf{88.6}}} & 92.1         & \textit{\textbf{86.5}}\\

\Xhline{3\arrayrulewidth}
\end{tabular}
}
}
\hfill
\parbox{.6\textwidth}{
\centering
\caption{\textbf{The attack success rate on ImageNet dataset.} U-SAGAN and U-BigGAN represent the base GAN models for U-GAN are SAGAN and BigGAN, respectively.}
\resizebox{0.6\textwidth}{!}{

\label{tab:imagenet}
\begin{tabular}{l|cccccc|c}
\Xhline{3\arrayrulewidth}  
\multirow{3}{*}{Method} & \multicolumn{6}{c|}{ASR(\%)}                                                         & \multirow{3}{*}{Time (s)} \\ \cline{2-7}
                        & \multicolumn{3}{c|}{ResNet50}                  & \multicolumn{3}{c|}{WideResNet50-2} &                           \\
                        & Clean & DiffPure & \multicolumn{1}{c|}{PGD-AT} & Clean     & DiffPure    & PGD-AT    &                           \\ \hline
AutoAttack              & 95.1  & 22.2     & \multicolumn{1}{c|}{56.2}   & 94.9      & 20.6        & 55.4      &   0.5                        \\ \hline
U-SAGAN                 & 99.3  & 30.5     & \multicolumn{1}{c|}{80.6}   & 98.9      & 28.6        & 70.1     &    10.4                       \\
U-BigGAN                & 96.8  & 40.1     & \multicolumn{1}{c|}{81.5}   & 96.5      & 35.5       & 78.4      &    11.2                       \\
AdvDiffuser             & 95.4  & 28.9     & \multicolumn{1}{c|}{90.6}   & 94.6      & 26.5           & 88.9      &        38.6                   \\
DiffAttack              & 92.8  & 30.6     & \multicolumn{1}{c|}{88.4}   & 90.6      & 27.6           & 85.3      &         28.2                  \\

AdvDiff                 & \textbf{99.8}  & 41.6     & \multicolumn{1}{c|}{92.4}   & \textbf{99.9}      & 38.5       & 90.6      &    \textbf{9.2}        \\

AdvDiff-Untargeted      & 99.5  & \textbf{75.2}     & \multicolumn{1}{c|}{\textbf{94.5}}   & 99.4      & \textbf{70.5}        & \textbf{92.6}      &    9.6                       \\

\Xhline{3\arrayrulewidth}                 
\end{tabular}
}
}
\end{table}

\noindent \textbf{ImageNet} It is reported that deep learning models on ImageNet are extremely vulnerable to adversarial attacks. However, the state-of-the-art adversarial defense DiffPure \cite{nie2022DiffPure} and adversarial training \cite{madry2017towards} can still defend against the perturbation-based attacks, as reported in Table \ref{tab:imagenet}. More UAEs evade the current defenses, but the generation quality of U-GAN is relatively poor compared to our adversarial examples. This phenomenon also shows that the performance of UAEs is heavily affected by the generation quality of the generation model. The adversarial examples generated by AdvDiff are more aggressive and stealthy than U-GAN's. Meanwhile, the generation speed of AdvDiff is the best among all the unrestricted adversarial attack methods. Note that we adopt the clean images generated by LDM to achieve DiffAttack and AutoAttack for a fair comparison.

\subsection{Generation Quality: True ASR for UAEs}

We witness similar ASR with U-GAN and AdvDiff. However, imperceptibility is also critical for a successful unrestricted adversarial attack, so we adopt the evaluation metrics in \cite{dhariwal2021diffusion} to compare the generation quality with and without performing unrestricted attacks. Table \ref{tab:gvd} shows that the AdvDiff achieves an overwhelming better IS score and similar FID score on the large-scale ImageNet dataset, where FID \cite{heusel2017gans} and IS \cite{salimans2016improved} scores are commonly adopted for evaluating the quality of a generative model. Because the generation of UAEs does not modify the data distribution of the generated images, the Precision score can be inferred as generation quality, while the Recall score indicates the flipped-label problems. We witness the frequent generation of flipped-label UAEs and low-quality UAEs from GAN-based methods, which is reflected by the decrease in the Precision score and the increase in the Recall score. Figure \ref{fig:gvd} illustrates this problem with some examples. It can be further proved that U-BigGAN achieves much higher image quality on non-reference metrics than reference metrics, as shown in Table \ref{tab:more}.

\begin{table}[t]
\caption{\textbf{The generation performance on the ImagetNet dataset.}}

\begin{center}
\resizebox{0.8\columnwidth}{!}{  
\begin{tabular}{l|ccc|cc}

\Xhline{3\arrayrulewidth}

Method             & FID ($\downarrow$)  & sFID ($\downarrow$) & IS ($\uparrow$)   & Precision ($\uparrow$)  & Recall ($\uparrow$)  \\ \hline
SAGAN    & 41.9      & 50.2    & 26.7      & 0.50      & 0.51   \\
BigGAN   & 19.3      & 45.7    & 250.3     & 0.95      & 0.21   \\
LDM      & 12.3      & 25.4    & 385.5     & 0.94      & 0.73   \\ \hline
U-SAGAN  & 52.8/+26\% & 52.2/+4\%  & 12.5/-53\%  & 0.58      & 0.57   \\
U-BigGAN & 25.4/+31\%   & 52.1/+14\% & 129.4/-48\% & 0.81      & 0.35   \\
AdvDiffuser & 26.8/+117\% & 38.6/+51\%  & 206.8/-46\%  & 0.70      & 0.75   \\
DiffAttack & 20.5/+66\%   & 40.2/+58\% & 264.3/-31\% & 0.83      & 0.73   \\
AdvDiff  & \textbf{16.2}/+31\%    & \textbf{30.4}/+20\% & \textbf{343.8}/-10\%  & \textbf{0.90}      & 0.75   \\
AdvDiff-Untargeted & 22.8/+85\%    & 33.4/+28\% & 220.8/-45\%  & 0.85      & \textbf{0.76}   \\
\Xhline{3\arrayrulewidth}  
\end{tabular}
}
\end{center}

\label{tab:gvd}
\end{table}

\begin{table}[t]
\caption{\textbf{The image quality on the ImagetNet dataset.}}

\begin{center}
\resizebox{0.8\columnwidth}{!}{  
\begin{tabular}{l|ccc|cc}

\Xhline{3\arrayrulewidth}

Method             & FID ($\downarrow$)  & LPIPS ($\downarrow$) & SSIM ($\uparrow$)   & BRISQUE\cite{mittal2011blind} ($\downarrow$)  & TRES ($\uparrow$)  \\ \hline
AutoAttack      & 26.5      & 0.72    &  0.21     & 34.4      & 69.8   \\ \hline
U-BigGAN & 25.4   & 0.50  & 0.32 & 19.4      & 80.3   \\
AdvDiffuser & 26.8   & 0.21  & 0.84 & 18.9      & 75.6   \\
DiffAttack & 20.5   & 0.15  & 0.75 & 22.6      & 67.8   \\
AdvDiff  & \textbf{16.2}    & \textbf{0.03} & \textbf{0.96}  & \textbf{18.1}      & \textbf{82.1}   \\
AdvDiff-Untargeted  & 22.8    & 0.14 & 0.85 & 16.2      & 76.8   \\
\Xhline{3\arrayrulewidth}  
\end{tabular}
}
\end{center}

\label{tab:more}
\end{table}

We find the IS score is heavily affected by the transferability of adversarial examples due to the calculation method. Therefore, we further compare the image quality of adversarial examples by commonly used metrics in Table \ref{tab:more}. The results show that AdvDiff (average 5 out of 5) and AdvDiff-Untargeted (average 4 out of 5) outperform existing adversarial attack methods using diffusion models. The perturbation-based adversarial attacks, i.e., AutoAttack, achieve much worse image quality compared with UAEs.

\subsection{UAEs against Defenses and Black-box Models}

Current defenses assume the adversarial examples are based on perturbations over data from the training dataset, i.e., $x_{adv} = x + \nabla\mathcal{L}, x \in D$. However, UAEs are synthetic data generated by the generative model. Because of different data sources, current defenses are hard to defend UAEs, which brings severe security concerns to deep learning applications. The proposed AdvDiff achieves an average of 36.8\% ASR against various defenses, while AutoAttack only achieves 30.7\% ASR with significantly lower image quality. We also test the attack transferability of AdvDiff and the results show that the untargeted version of AdvDiff achieves the best performance against black-box models. Experiment results are given in Table \ref{tab:db}.

\begin{table}[t]

\caption{\textbf{The attack success rates (\%) of ResNet50 examples for transfer attack and attack against defenses on the ImagetNet dataset.}}

\begin{center}
\resizebox{0.85\columnwidth}{!}{  
\begin{tabular}{l|cccc}

\Xhline{3\arrayrulewidth}
Method        & ResNet-152 \cite{he2016deep} & Inception v3 \cite{szegedy2016rethinking} &  ViT-B \cite{dosovitskiy2020image} &  BEiT \cite{bao2021beit}  \\ \hline
AutoAttack        & 32.5             & 38.6 & 9.3      & 45.3          \\ \hline
U-BigGAN          & 30.8             & 35.3                & 30.1      & 69.4        \\
AdvDiffuser         & 18.3             & 20.0  & 18.5      & 79.4     \\ 
DiffAttack         & 21.1             & \textbf{43.9}  & 17.4      & 78.0    \\ 
AdvDiff          & 20.5              & 14.9                &  17.8       & 78.8    \\ 
AdvDiff-Untargeted          & \textbf{52.0}              &  42.7                & \textbf{36.0}     & \textbf{81.5}       \\ \Xhline{3\arrayrulewidth}
Method        & Adv-Inception \cite{madry2017towards} & AdvProp \cite{xie2020adversarial} & DiffPure \cite{nie2022DiffPure} & HGD \cite{liao2018defense} \\ \hline
AutoAttack       & 14.6               & 69.6  & 22.2      & 20.5       \\ \hline
U-BigGAN          & 40.6              & 75.2                & 40.1 &  22.6         \\
AdvDiffuser          & 24.4              & 84.0                & 30.5 & 10.8             \\
DiffAttack          & 30.9              & 85.1               & 30.6 & 20.5             \\
AdvDiff          & 19.4            & 89.7              & 41.6 & 17.8        \\ 
AdvDiff-Untargeted          & \textbf{60.1}              & \textbf{95.3}              & \textbf{75.2} & \textbf{53.8}        \\ \Xhline{3\arrayrulewidth}
Method        & R\&P \cite{xie2017mitigating} & RS \cite{cohen2019certified} & NRP \cite{naseer2020self} & Bit-Red \cite{xu2017feature} \\ \hline
AutoAttack       & 20.6               & 38.9  & 39.4     &   19.8     \\ \hline
U-BigGAN          & 14.2              & 34.5                & 30.9     & 13.1     \\
AdvDiffuser          & 15.4              & 38.4                & 40.5  & 11.4             \\
DiffAttack          & 23.7              & 40.8                & 38.5  &  20.1           \\
AdvDiff          & 17.4             & 47.6              & 45.2 & 15.8        \\
AdvDiff-Untargeted          & \textbf{56.8}             & \textbf{82.8}              & \textbf{74.2} & \textbf{52.6}        \\ \Xhline{3\arrayrulewidth}
\end{tabular}
}
\end{center}
\label{tab:db}

\end{table}

\subsection{Better Adversarial Diffusion Sampling}

We present detailed comparisons with DiffAttack and AdvDiffuser. The results show that the proposed adversarial guidance achieves significantly higher generation quality than PGD-based adversarial guidance. With PGD gradient guidance, the diffusion model generates images with a similar Recall score but a much lower Precision score, which indicates that the PGD gradient influences the benign generation process and causes the generation of low-quality images.
The result proves that the adversarial guidance of diffusion models should be carefully designed without affecting the benign sampling process. Meanwhile, the generation speed of AdvDiff is the best among the existing diffusion attack methods.  Note that AdvDiff (36.8\%) sightly outperforms AdvDiffuser (32.0\%) and DiffAttack (36.2\%) against defenses. 
 However, previous attacks achieve slightly better transfer attack performance than the original AdvDiff. The reason could be the gradient of the cross-entropy loss is shared among nearly all the deep learning models and is better at attack transferability against these models. Nevertheless, the untargeted version of AdvDiff achieves overwhelmingly better performance, which further demonstrates the effectiveness of the proposed adversarial sampling. But the generation quality is affected, we leave a better design in the future work.

\begin{figure}[t]
   \begin{center}
     \includegraphics[width=1.0\linewidth]{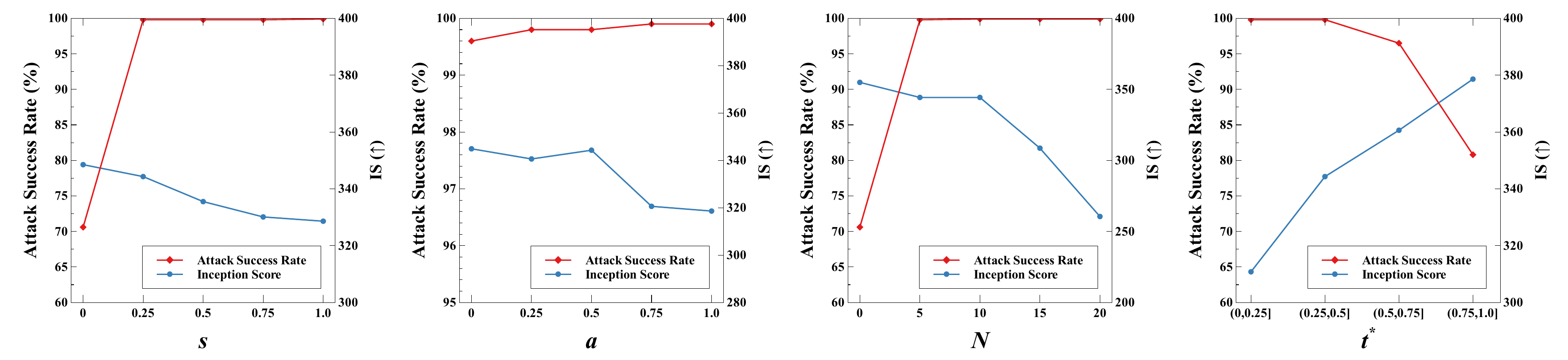}
   \end{center}

      \caption{\textbf{Ablation study of the impact of parameters in AdvDiff.} The results are generated from the ImageNet dataset against the ResNet50 model. We adopt the ASR and IS scores to show the impact of attack performance and generation quality.}
   \label{fig:abl}

   \end{figure}
   
\subsection{Ablation Study}

We discuss the impact of the parameters of AdvDiff in the subsection. Note that our proposed method does not require re-training the conditional diffusion models. The ablation study is performed only on the sampling process.

\noindent \textbf{Adversarial guidance scale $s$ and $a$}. The magnitudes of $s$ and $a$ greatly affect the ASR of AdvDiff, as shown in Figure \ref{fig:abl}. Noted that we witness the generation of unrealistic images when setting the adversarial guidance extremely large. See the Appendix for detailed discussions.

\noindent \textbf{Noise sampling guidance steps $N$}.  Like the iteration times of GAN-based unrestricted adversarial attacks, larger steps $N$ can effectively increase the attack performance against an accurate classifier, as shown in Figure \ref{fig:abl}. However, it can affect the initial noise distribution and hence decreases the generation quality. During experiments, we observe that adversarial guidance is already capable of generating adversarial examples with high ASR. Thus, we can set a small noise sampling guidance step $N$ for better sample quality.

\noindent \textbf{Adversarial guidance timestep $t^*$}. The reverse diffusion process gradually denoises the input noise. Therefore we generally get noisy images at most timesteps. Because the target classifier is not able to classify the noisy input, the adversarial guidance is not effective in the early reverse diffusion process. Figure \ref{fig:abl} shows our results, and we can improve the performance of adversarial guidance by training a separate classifier, which we leave for future work.

\section{Conclusion}

In this work, we propose a new method called AdvDiff, which can conduct unrestricted adversarial attacks using any pre-trained conditional diffusion model. We propose two novel adversarial guidance techniques in AdvDiff that lead diffusion models to obtain high-quality, realistic adversarial examples without disrupting the diffusion process. Experiments show that our AdvDiff vastly outperforms GAN-based and diffusion-based attacks in terms of attack success rate and image generation quality, especially in the ImageNet dataset. AdvDiff indicates that diffusion models have demonstrated effectiveness in adversarial attacks, and highlights the need for further research to enhance AI model robustness against unrestricted attacks.

\section*{Acknowledgements}
This work was supported in part by HK RGC GRF under Grant PolyU 15201323.

\bibliographystyle{splncs04}
\bibliography{main}
\end{document}


\title{Supplementary Materials for \\
AdvDiff: Generating Unrestricted Adversarial Examples using Diffusion Models} 

\titlerunning{AdvDiff}

\author{Xuelong Dai\inst{1}\orcidlink{0000-0001-6646-6514} \and
Kaisheng Liang\inst{1}\orcidlink{0000-0002-8297-6378} \and
Bin Xiao\inst{1}\orcidlink{0000-0003-4223-8220}}

\authorrunning{X.~Dai et al.}

\institute{
The Hong Kong Polytechnic University\\
\email{\{xuelong.dai, kaisheng.liang\}@connect.polyu.hk, b.xiao@polyu.edu.hk }}

\maketitle


\setcounter{equation}{10}
\setcounter{algorithm}{1}

\section{Detailed Proof of Equation 8}

 We can obtain the sample $x_{t-1}$ with condition label $y$, according to the sampling with the classifier-free guidance. To get the unrestricted adversarial example $x^*_{t-1}$, we add adversarial guidance to the conditional sampling process with Equation 8. With Bayes' theorem, we want to deduce the adversarial sampling with adversarial guidance at timestep $t$ by: 
\begin{equation}
    p({x}^*_{t-1}|y_{a}) = \frac{p(y_{a}|{x}^*_{t-1})p({x}^*_{t-1})}{p(y_{a})}
    \label{eq:s1}
\end{equation}
with Equation \ref{eq:s1}, we want to sample the adversarial examples with the target label $y_a$. Starting from $x_t$, the sampling of the reverse generation process with AdvDiff is:
\begin{align}
    p({x}^*_{t-1}|x_{t},y_{a}) =& \frac{p(y_{a}|{x}^*_{t-1},x_{t})p({x}^*_{t-1}|x_{t})}{p(y_{a}|x_{t})} 
    \label{eq:s2}
\end{align}
Noted that Equation \ref{eq:s2} is the same as the deviation of classifier-guidance in \cite{dhariwal2021diffusion}'s Section 4.1, where they treated $p(y_{a}|x_{t})$ as a constant. Because $p({x}^*_{t-1}|x_{t})$ is the known sampling process by our conditional diffusion sampling, we evaluate $\frac{p(y_{a}|{x}^*_{t-1},x_{t})}{p(y_{a}|x_{t})}$ by:
\begin{equation}
    \log p_f(y_{a}|{x}^*_{t-1}) - \log p_f(y_{a}|x_{t})
    \label{eq:s3}
\end{equation}
We can approximate Equation \ref{eq:s3} using a Taylor expansion around ${x}^*_{t-1} = \mu(x_t)$ as:
\begin{align}
    \log p_f(y_{a}|{x}^*_{t-1}) - \log p_f(y_{a}|x_{t}) &\approx \log p_f(y_{a}|\mu(x_{t})) \notag\\ &+  ({x}^*_{t-1}-\mu(x_{t}))\nabla_{\mu(x_{t})} \log p_{f}({y_{a}}|\mu(x_{t})) \notag\\ &- \log p_f(y_{a}|x_{t}) + C  \notag \\ &= ({x}^*_{t-1}-\mu(x_{t}))\nabla_{\mu(x_{t})} \log p_{f}({y_{a}}|\mu(x_{t}))+C
\end{align}
Assume $p({x}^*_{t-1}|x_{t}) = \mathcal{N}(x^*_{t-1};\mu(x_t),  \sigma_t^2\textbf{I}) \propto e^{{-(x^*_{t-1}-\mu(x_t))^2}/{2\sigma_t^2}}$, we have:
\begin{align}
   p({x}^*_{t-1}|x_{t},y_{a}) &\propto e^{{-(x^*_{t-1}-\mu(x_t))^2}/{2\sigma_t^2} +({x}^*_{t-1}-\mu(x_{t}))\nabla_{\mu(x_{t})} \log p_{f}({y_{a}}|\mu(x_{t}))} \notag  \\ &\propto e^{{-(x^*_{t-1}-\mu(x_t)-\sigma_t^2\nabla_{\mu(x_{t})} \log p_{f}({y_{a}}|\mu(x_{t})))^2}/{2\sigma_t^2}+(\nabla_{\mu(x_{t})} \log p_{f}({y_{a}}|\mu(x_{t})))^2/{2\sigma_t^2}} \notag \\&\propto  e^{{-(x^*_{t-1}-\mu(x_t)-\sigma_t^2\nabla_{\mu(x_{t})} \log p_{f}({y_{a}}|\mu(x_{t})))^2}/{2\sigma_t^2}+C} \notag \\ &\approx \mathcal{N}(x^*_{t-1};\mu(x_t)+\sigma_t^2\nabla_{\mu(x_{t})} \log p_{f}({y_{a}}|\mu(x_{t})),  \sigma_t^2\textbf{I} )
   \label{eq:s4}
\end{align}
Sampling with Equation \ref{eq:s4} should be:
\begin{equation}
    {x}^*_{t-1} = \mu({x}_t,y) + \sigma_t{\varepsilon} +\sigma_t^2 s\nabla_{\mu(x_{t})} \log p_{f}({y_{a}}|\mu(x_{t}))
\end{equation}
where $\mu({x}_t,y)$ is the conditional mean value and $\varepsilon$ is sampled from ${\varepsilon}\sim \mathcal{N}(0,\textbf{I})$. Note that $ \mu({x}_t,y) + \sigma_t{\varepsilon}$ is the normal sampling process that we will get $x_{t-1}$. In practice, in each diffusion step, the difference between $x_{t-1}$ and $\mu(x_t)$ should be small enough \cite{dhariwal2021diffusion,ho2020denoising} for a reasonable and stable diffusion sampling. Therefore, we adopt $x_{t-1}$ to calculate the adversarial gradient after the sampling with the conditional diffusion model, and we have:
\begin{equation}
    {x}^*_{t-1} = \mu({x}_t,y) + \sigma_t{\varepsilon} +\sigma_t^2 s\nabla_{\mu(x_{t})} \log p_{f}({y_{a}}|\mu(x_{t})) \approx x_{t-1}+\sigma_t^2 s\nabla_{{x}_{t-1}} \log p_{f}({y_{a}}|{x}_{t-1})
\end{equation}
where $s$ is the adversarial guidance scale. \hfill $\square$

\section{Detailed Proof of Equation 10}
The deviation of Equation 10 is similar to Equation 8, where the noise sampling guidance is added with the forward diffusion process. Similarly, we have Equation 9:
\begin{align}
    p({x}_{T}|y_{a}) =& \frac{p(y_{a}|{x}_{T})p({x}_{T})}{p(y_{a})} = \frac{p(y_{a}|x_T,x_0)p(x_T|x_0)}{p(y_{a}|x_0)} 
    \label{eq:noise}
\end{align}
And Taylor expansion around $x_{T}=x_0$ to evaluate $\frac{p(y_{a}|x_T,x_0)}{p(y_{a}|x_0)}$.
\begin{align}
    \log p_f(y_{a}|x_T) - \log p_f(y_{a}|x_0) = (x_T-x_0)\nabla_{x_0} \log p_{f}({y_{a}}|x_0)+C
\end{align}
From $x_0$ to $x_T$, we gradually add the Gaussian noise with the predefined schedule \cite{ho2020denoising}:
\begin{equation}
    p(x_T|x_0)=\mathcal{N}(x_T;\sqrt{\bar{\alpha}_T}x_0,(1-\bar{\alpha}_T)\textbf{I})
\end{equation}
The noise sampling guidance is as follows:
\begin{align}
    {x}_{T} &\approx (\bar\mu({x}_0,y) + \bar{\sigma}_T{\varepsilon}) +\bar{\sigma}_T^2 a\nabla_{x_0} \log p_{f}({y_{a}}|x_0) \notag \\&= x_T +\bar{\sigma}_T^2 a\nabla_{{x}_0} \log p_{f}({y_{a}}|{x}_0)
\end{align}
where $\bar\mu({x}_0,y) + \bar{\sigma}_T{\varepsilon}$ is the forward diffusion process to get $x_T$ with $x_0$ and $a$ is the  noise sampling guidance scale.  \hfill $\square$

\section{AdvDiff for DDIM}
We give the derivation for AdvDiff for DDIM followed with \cite{dhariwal2021diffusion}. The score function for the DDIM diffusion model is:
\begin{equation}
    \nabla_{ {x}}\log p_f( {x}| {y}) = \nabla_{ {x}}\log p_f( {x}) + \nabla_{ {x}}\log p_f( {y}| {x})
\end{equation}
We set $y$ as our adversarial guidance $y_a$:
\begin{align}
    \nabla_{ {x}}\log p_f( {x}| {y}_a) &= \nabla_{ {x}}\log p_f( {x}) + \nabla_{ {x}}\log p_f( {y}_a| {x}) \notag \\ &= -\frac{1}{\sqrt{1-\bar{\alpha}}}\epsilon_\theta(x) + \nabla_{ {x}}\log p_f( {y}_a| {x})
\end{align}
Finally, the new epsilon prediction $\hat{\epsilon}_\theta(x_t)$ is defined as follows:
\begin{equation}
    \hat{\epsilon}_\theta(x_t) = \epsilon_\theta(x_t) - \sqrt{1-\bar{\alpha}_t}\nabla_{ {x}_t}\log p_f( {y}_a| {x}_t)
\end{equation}
Then the DDIM with AdvDiff is Algorithm \ref{alg:s1} over the trained classifier-free diffusion model $\epsilon_{\theta}(\cdot)$.
\begin{algorithm}[tb]
  \caption{DDIM Adversarial Diffusion Sampling} 
  \label{alg:s1}
  \begin{algorithmic}[1]
    \Require $y_a$: target label for adversarial attack
    \Require $y$: ground truth class label
    \Require $s,a$: adversarial guidance scale
    \Require $w$: classification guidance scale
    \Require $N$: noise sampling guidance steps
    \Require $T$: reverse generation process timestep

    \State $x_{T} \sim \mathcal{N}(0, \textbf{I})$
    \State $x_{adv} = \varnothing $
    \For{$i = 1\ldots N$}
    \For{$t=T, \dotsc, 1$}

      \State $\tilde{\epsilon}_t = (1+w)\epsilon_\theta(x_{t}, y) - w\epsilon_{\theta}(x_{t})$ 
      \State     $\hat{\epsilon}_t = \tilde{\epsilon}_t - \sqrt{1-\bar{\alpha}_t}\nabla_{ {x}_t}\log p_f( {y}_a| {x_t})$
      \State Classifier-free DDIM sampling $x_{t-1}$ with $\hat{\epsilon}_t$
    \EndFor
      \State Obtain classification result from $f(x_0)$
      \State Compute the gradient with $\log p_{f}({y_{a}}|{x}_0)$
      \State Update $x_{T}$ by $x_T = x_T + a\nabla_{{x}_0} \log p_{f}({y_{a}}|{x}_0)$
      \State  $x_{adv} \gets x_{0}$ if $f(x_0)=y_a$

    \EndFor
    \State \textbf{return} $x_{adv}$
  \end{algorithmic}
\end{algorithm}

We can further deduce the DDIM with $\hat{\epsilon}_\theta(x_t)$ by:
\begin{align}
  x_{t-1}&= \sqrt{\bar{\alpha}_{t-1}} \left( \frac{x_t - \sqrt{1-\bar{\alpha}_t} \hat{\epsilon}_\theta}{\sqrt{\bar{\alpha}_t}} \right) + \sqrt{1-\bar{\alpha}_{t-1}} \hat{\epsilon}_\theta \notag \\&=\sqrt{\bar{\alpha}_{t-1}} \left( \frac{x_t - \sqrt{1-\bar{\alpha}_t}{\epsilon}_\theta}{\sqrt{\bar{\alpha}_t}} \right) + \sqrt{1-\bar{\alpha}_{t-1}} {\epsilon}_\theta+C\cdot\nabla_{ {x}_t}\log p_f( {y}_a| {x}_t)
  \label{eq:s5}
\end{align}
where we can replace $C$ with our adversarial guidance scale.

\section{Related Work}

Since Szegedy et al. \cite{szegedy2013intriguing} had proved that DL models are extremely vulnerable to adversarial attacks, researchers have been digging into improving the model's adversarial robustness by proposing stronger adversarial attack methods and their counter-measurements. 

\textbf{Perturbation-based adversarial examples with generative models}: Most related works performed adversarial attacks by perturbing a subset of clean data to fool the target classifier. These attacks \cite{carlini2017towards,madry2017towards,kurakin2018adversarial} attempted to generate better perturbations with higher attack success rates and smaller perturbations. With the emergence of generative models, end-to-end adversarial attacks \cite{baluja2018learning,xiao2018generating,poursaeed2018generative} have greatly improved the generation efficiency by pre-training the generative module. These methods integrate the advertorial loss into the training of generative models and generate adversarial examples by trained generators with clean data.

\textbf{Unrestricted adversarial examples with generative models}: Perturbation-based adversarial examples require insignificant norm distance to the given clean data in order to guarantee the indistinguishability, which only covers a small fraction of all possible adversarial examples \cite{song2018constructing}. To remove such restrictions, Song et al. \cite{song2018constructing} proposed an unrestricted adversarial attack method that searches over the latent space of the input noise vector with an adversarial loss function and a well-trained AC-GAN \cite{odena2017conditional}. Inspired by Song's work \cite{song2018constructing}, recent works \cite{poursaeed2019fine,xiang2022egm} made improvements in the generation quality and generation efficiency of UAEs. Diffusion model based adversarial attacks \cite{chen2023advdiffuser,chen2023diffusion,chen2023content} also achieve satisfying attack performance against deep learning models. However, the performance of existing approaches suffers from the unstable training of GAN models as well as the lack of theoretical support for injecting PGD-based gradients. Therefore, we provide an effective and theoretically analyzed solution with the diffusion model in this paper. 

\subsection{Conditional Diffusion Model for Image Generation}

Diffusion models have shown great generation quality and diversity in the image synthesis task since Ho et al.\cite{ho2020denoising} proposed a probabilistic diffusion model for image generation that greatly improved the performance of diffusion models.  Diffusion models for conditional image generation are extensively developed for more usable and flexible image synthesis. Dhariwal \& Nichol \cite{dhariwal2021diffusion} proposed a conditional diffusion model that adopted classifier-guidance for incorporating label information into the diffusion model. They separately trained an additional classifier and utilized the gradient of the classifier for conditional image generation. Jonathan Ho \& Tim Salimans \cite{ho2021classifier} performed the conditional guidance without an extra classifier to a diffusion model. They trained a conditional diffusion model together with a standard diffusion model. During sampling, they adopted the combination of these two models for image generation. Their idea is motivated by an implicit classifier with the Bayes rule. Followed by \cite{dhariwal2021diffusion,ho2021classifier}'s works, many research \cite{rombach2022high,nichol2021glide,lugmayr2022repaint,gafni2022make} have been proposed to achieve state-of-the-art performance on image generation, image inpainting, and text-to-image generation tasks. Despite utilizing diffusion models for image generation has been widely discussed, none of these works have discovered the adversarial examples generation method with the diffusion model. Also, it is a new challenge to defend against the adversarial examples generated by the diffusion model.

\section{Implementation Details}
As AdvDiff supports both DDPM and DDIM sampling, we adopt LDM
\footnote{\url{https://github.com/CompVis/latent-diffusion}} with DDIM sampler for the experiment on ImageNet for reproducibility and poor performance of simple DDPM on ImageNet. We adopt 500 sampling steps for DDPM on MNIST and 200 sampling steps for LDM on ImageNet. For conditional sampling, we use one-hot label information for both DDPM and DDIM sampling for a fair comparison with GAN. The noise sampling step is set as $(0,0.5]$ for the MNIST dataset and $(0,0.2]$ for the ImageNet dataset. We follow the default settings in DiffAttack and AdvDiffuser in the experiments.

\section{More Experiment Results}

\begin{figure}
   \begin{center}
     \includegraphics[width=0.8\linewidth]{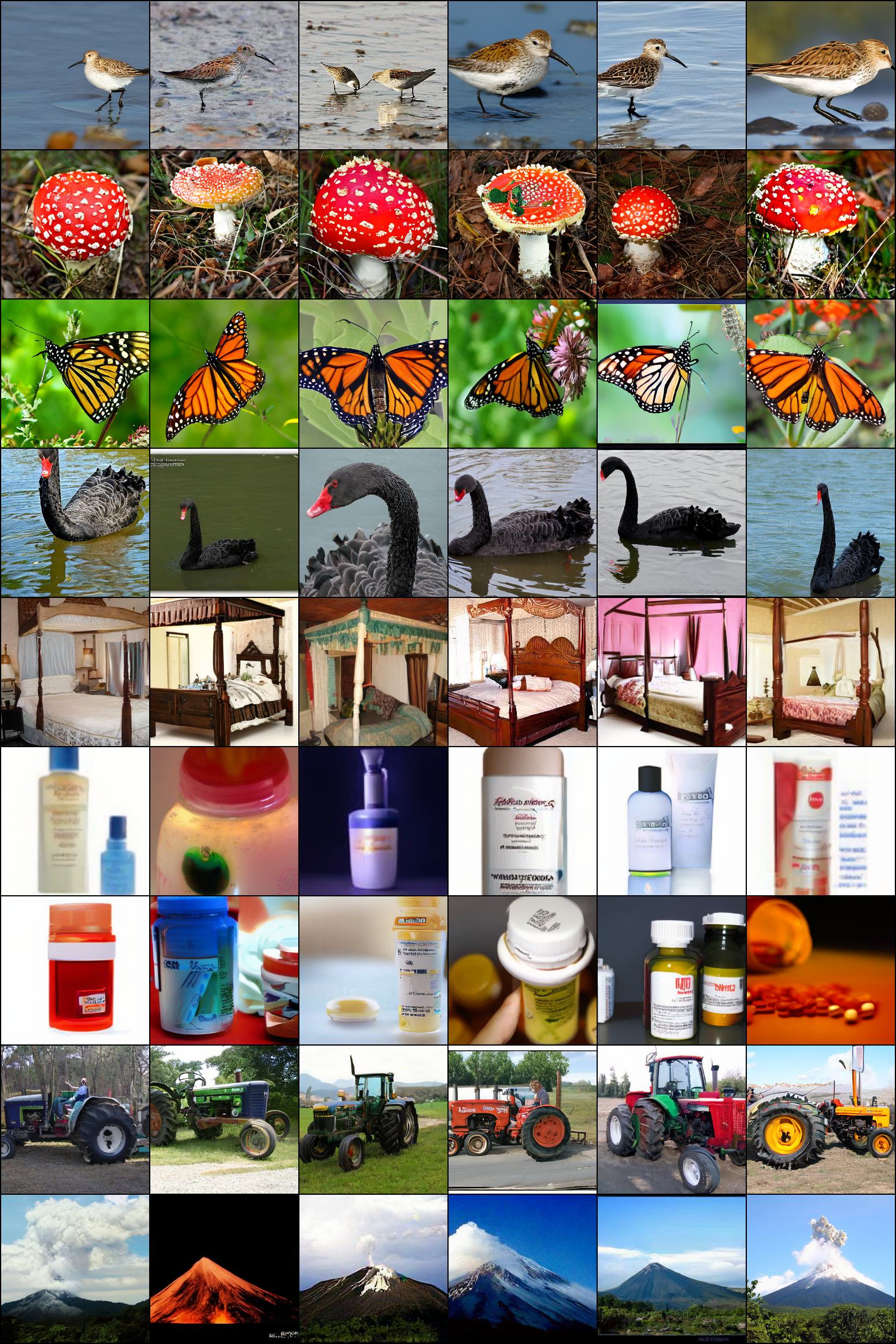}
   \end{center}
      \caption{\textbf{More unrestricted adversarial examples generated by AdvDiff on the ImageNet dataset.} }
   \label{fig:s1}
   \end{figure}

\begin{figure}
   \begin{center}
     \includegraphics[width=0.5\linewidth]{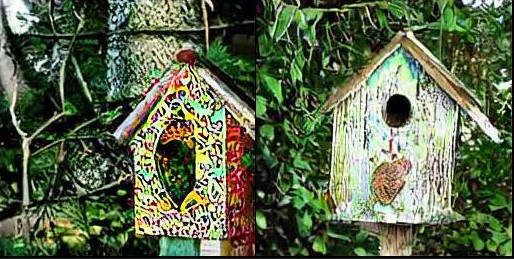}
   \end{center}
      \caption{\textbf{Failure cases when $s=10.0$.}}
   \label{fig:s2}
   \end{figure}

\begin{figure}
   \begin{center}
     \includegraphics[width=0.5\linewidth]{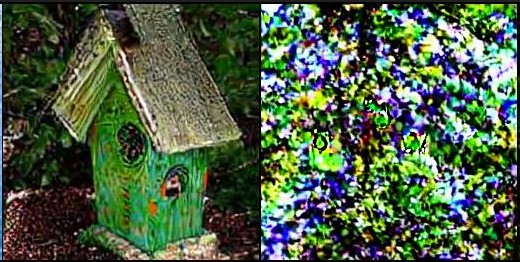}
   \end{center}
      \caption{\textbf{Failure cases when $a=10.0$.} }
   \label{fig:s3}
   \end{figure}

We give more experiment results in Figure \ref{fig:s1} to demonstrate the generation quality on the ImageNet dataset. We also provide some failure cases of our AdvDiff, which happens when we set the adversarial guidance scale $s$ and $a$ extremely large. Figure \ref{fig:s2} shows that a large $s$ (10.0) tends to generate images with noisy textures while a large $a$ (10.0) can generate noisy images. Figure \ref{fig:s3} shows that modifying the initial noise with $a$ can disturb the noise distribution if we add the gradient in an irrational manner.

   \begin{figure}
   \begin{center}
     \includegraphics[width=0.7\linewidth]{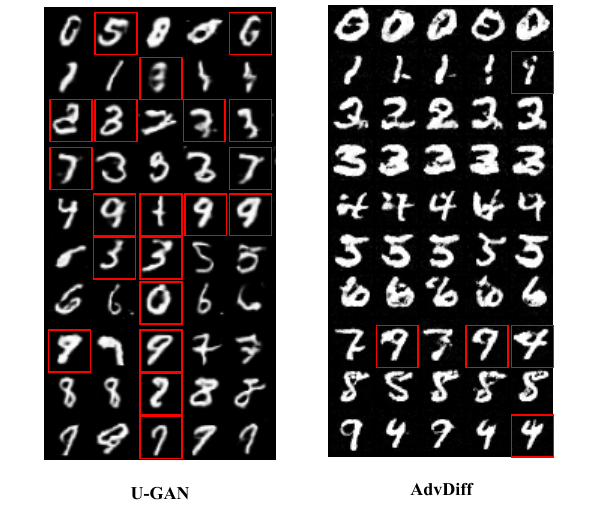}
   \end{center}
      \caption{\textbf{User study on MNIST datast.} Flipped-label UAEs are tagged with a red box. MNIST dataset is robust against UAEs because each image only contains $28\times28$ pixels. }
   \label{fig:us}
   \end{figure}
   
\section{AdvDiff against Adversarial Training with Diffusion Models}

\begin{table}[ht]

\begin{center}

\caption{\textbf{Performance under AdvDiff attack against adversarial training on the ResNet18 model.}}
\label{tab:at}
\resizebox{1.0\columnwidth}{!}{  
\begin{tabular}{c|ccc|ccc|ccc}
\hline
Method   & Clean & PGD & AdvDiff Attack & AT-UAE  & PGD  & AdvDiff Attack & AT-PGD  & PGD  & AdvDiff Attack \\ \hline
Accuracy (\%) & 99.0 & 0.7  & 7.9            & 99.2 & 16.8 & 32.6  & 95.2 & 79.2 & 13.5     \\
\hline   
\end{tabular}
}
\end{center}
\end{table}

Adversarial training is an effective way to improve classification accuracy against adversarial attacks. Thus, it should be an effective way to defend against UAEs. However, UAEs are generated from random noise latents rather than fixed gradient perturbations by given input images. Therefore, adversarial training with UAEs is not as effective as it is with perturbation-based attacks. We test the AT-UAE with UAEs generated by AdvDiff on the MNIST dataset with 1000 images per class. The results are given in Table \ref{tab:at}. The result shows that AT with UAEs improves the robust accuracy against AdvDiff, but the performance is limited as there is an infinite number of random latents to generate UAEs.

\section{Improving Attack Transferability}

AdvDiff achieves overwhelmingly better generation quality and attack success rate against white-box target models by adversarial diffusion sampling with a given target label $y$. However, the attack transferability is limited due to different decision boundaries from black-box models. Normally, black-box attackers use the gradient of the original label to generate perturbations. Therefore, we adopt the same settings to improve the attack transferability of AdvDiff (denoted as AdvDiff-Untargeted), i.e., $- \nabla_{{x}_{t-1}} \log p_{f}(y|{x}_{t-1})$, where $y$ is the ground truth label to generate samples. However, such sampling will decrease the generation quality as sampling from the negative distribution does not follow the benign diffusion process. Table \ref{tab:dbs} shows that the attack transferability significantly improved with a decrease in generation quality.  We leave a better design of attack transferability for future work. Additional experiments against transformers are also given in Table \ref{tab:dbs}.

\begin{table}[t]

\caption{\textbf{The attack success rates (\%) of ResNet50 examples for transfer attack and attack against defenses on the ImagetNet dataset.}}

\vspace{-0.2in}
\begin{center}
\resizebox{\columnwidth}{!}{  
\begin{tabular}{c|ccc}

\Xhline{3\arrayrulewidth}
Method        & ResNet-152 \cite{he2016deep} & Inception v3 \cite{szegedy2016rethinking} &  DenseNet-121 \cite{huang2017densely}  \\ \hline
AutoAttack        & 32.5              & 38.6                 & 43.8               \\ \hline
U-BigGAN          & 30.8              & 35.3                & 16.8             \\
AdvDiffuser          & 18.3             & 20.0               & 24.8           \\
DiffAttack          & 21.1              & 43.9              & 23.8           \\
AdvDiff          & 20.5              & 14.9                &  35.8          \\ 
AdvDiff-Untargeted          & \textbf{52.0}              & \textbf{42.7}                &  \textbf{60.9}        \\ \Xhline{3\arrayrulewidth}
Method        & MobileNet v2 \cite{sandler2018mobilenetv2} & PNASNet \cite{liu2018progressive} &  MNASNet \cite{tan2019mnasnet}\\ \hline
AutoAttack       & 41.6              & 38.5  & 42.5             \\ \hline
U-BigGAN          & 18.4              & 22.1                & 16.8             \\
AdvDiffuser          & 30.3              & 15.2               & 26.7           \\
DiffAttack          & 22.3              & 26.9               & 30.4           \\
AdvDiff          & 15.4             & 23.2               & 38.9       \\
AdvDiff-Untargeted          & \textbf{49.5}             & \textbf{53.0}               & \textbf{47.6}        \\ \Xhline{3\arrayrulewidth}
Method        & VGG-19 \cite{simonyan2014very} & SENet \cite{hu2018squeeze} &  WRN \cite{zagoruyko2016paying}\\ \hline
AutoAttack       & 48.3               & 23.7               & 29.5            \\ \hline
U-BigGAN          & 18.4              & 22.1                & 16.8             \\
AdvDiffuser          & 28.7              & 18.8               & 22.0           \\
DiffAttack          & 30.0              & 22.1               & 23.6           \\
AdvDiff          & 16.8             & 10.0               & 11.8        \\
AdvDiff$_\text{transfer}$         & \textbf{58.5}             & \textbf{51.2}               & \textbf{57.4}        \\ \Xhline{3\arrayrulewidth}
Method        & ViT-B \cite{dosovitskiy2020image} & DeiT-B \cite{xie2020adversarial} & BEiT \cite{bao2021beit} \\ \hline
AutoAttack       & 9.3              & 8.9  & 45.3            \\ \hline
U-BigGAN          & 30.1              & 27.7                & 69.4             \\
AdvDiffuser          & 18.5              & 12.5               & 79.4           \\
DiffAttack          & 17.4              & 17.5               & 38.6           \\
AdvDiff          & 17.8              & 17.6              & 78.8        \\
AdvDiff-Untargeted          & \textbf{36.0}             & \textbf{58.5}               & \textbf{81.5}        \\ 
\Xhline{3\arrayrulewidth}
\end{tabular}
}
\end{center}
\label{tab:dbs}

\vspace{-0.2in}
\end{table}

\section{User Study}

\begin{table}[t]

\begin{center}

\caption{\textbf{User Study about flipped label problem on MNIST.}}
\label{tab:us}
\resizebox{0.5\columnwidth}{!}{  
\begin{tabular}{c|cc}
\hline
 Method & U-GAN & AdvDiff \\ \hline
User Study & 425/1000         & \textbf{102/1000}         \\ \hline
\end{tabular}
}
\end{center}
   
\end{table}

We further perform a user study to justify the performance of AdvDiff, where we ask 20 participants to identify flipped label images on the MNIST dataset with 5 images on each class by U-GAN and AdvDiff. The results are given in Table \ref{tab:us}. We also give the tagged examples on UAEs generated by U-GAN and AdvDiff in Figure \ref{fig:us}, where AdvDiff's UAEs are remarkably better in generation quality and harder to identify flipped label images than U-GAN.

\section{Improving the Generation Quality}

\begin{figure}
   \begin{center}
     \includegraphics[width=1.0\linewidth]{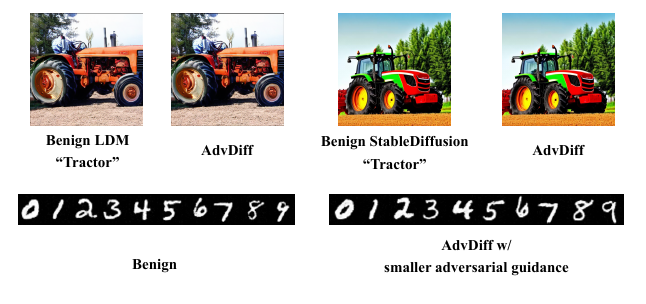}
   \end{center}
   
      \caption{The improvements for better generation quality.}
      
   \label{fig:imq}
   \end{figure}

AdvDiff crafts adversarial examples with imperceptible perturbations, making the generation quality of our methods largely reliant on the benign diffusion model's performance. Figure \ref{fig:imq} shows that AdvDiff produces higher-quality images when using StableDiffusion as the benign diffusion model.  Moreover, we can set the adversarial guidance to a smaller value for better quality with a decrease in the generation speed.  The guidance in the paper on the MNIST dataset aims at high ASR per batch for a fair comparison with previous attacks, while the visual quality can be affected by its limited $28\times28$ grey pixel space. We can also achieve stable AE generation by using latents obtained by conducting the forward diffusion process from the training dataset's clean images. 

\section{Comparing with Existing Diffusion Model Attacks}

\begin{figure}

   \begin{center}
     \includegraphics[width=1.0\linewidth]{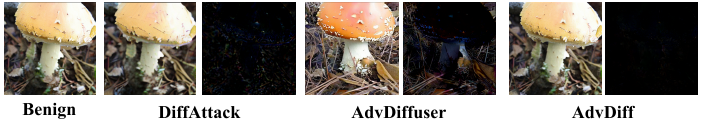}
   \end{center}
   
      \caption{The generated adversarial examples (mushroom) from different diffusion-based attacks and corresponding perturbations.}
      
   \label{fig:diffcomp}
   \end{figure}

There are several diffusion model adversarial attacks \cite{chen2023advdiffuser,chen2023diffusion,chen2023content} achieve state-of-the-art performance. However, most of them did not release the official code which makes it difficult to compare with these methods. All these works adopt the optimization over given loss functions (i.e., PGD-like gradient) to generate UAEs with the diffusion models. Figure \ref{fig:diffcomp} provides a direct comparison of adversarial examples from different methods. Our findings indicate that PGD-like adversarial guidance perturbations significantly alter the texture of benign images from AdvDiffuser. Similarly, the perturbations from DiffAttack are also very similar to standard PGD perturbations, where the perturbations are uniformly applied across the entire image. In contrast, our perturbations mainly target the mushroom's contour and are substantially less noticeable than those from existing attacks.

Our work can easily be combined with some exciting works by replacing the gradient with AdvDiff's adversarial guidance (especially for AdvDiffuser \cite{chen2023advdiffuser} which directly adopts the PGD gradient to conduct the adversarial attack). We hope our work can gain new insight for designing adversarial attacks using diffusion models.

\section{Comparing with 2021 CVPR Competition Winner}

We compare with the 1st winner \cite{liu2023towards} of 2021 CVPR unrestricted adversarial attack competition \cite{chen2021unrestricted} follows their official implementation on ImageNet. Two variants of \cite{liu2023towards}'s attacks are compared, which are GA-IFGSM and GA-FSA. The results are given in Table \ref{tab:cvpr}. The proposed AdvDiff outperforms \cite{liu2023towards}'s attack in terms of generation quality and attack performance. It may not be a fair comparison as \cite{liu2023towards}'s attack is not a synthetic attack.

\begin{table}[ht]
\caption{\textbf{The attack performance on the ImagetNet dataset.}}

\vspace{-0.2in}
\begin{center}
\resizebox{\columnwidth}{!}{  
\begin{tabular}{c|cc|ccccc}
\Xhline{3\arrayrulewidth} 
            & ASR           & PGD-AT        & FID           & LPIPS         & SSIM          & BRISQUE       & TRES          \\ \hline
AdvDiff-Untargeted & 99.5          & \textbf{94.5}          & 22.8          & 0.14          & 0.85          & 16.2          & 76.8          \\
AdvDiff     & 99.8 & 92.4 & \textbf{16.2} & \textbf{0.03} & \textbf{0.96} & \textbf{18.1} & \textbf{82.1} \\
GA-IFGSM    & 99.8          & 82.6          & 50.4          & 0.24          & 0.78          & 40.4          & 62.0          \\
GA-FSA      & \textbf{99.9} & 91.4          & 70.6          & 0.32          & 0.56          & 50.8          & 58.4     \\
\Xhline{3\arrayrulewidth}
\end{tabular}
}
\end{center}

\vspace{-0.2in}
\label{tab:cvpr}
\end{table}

\section{Discussion about perturbations, flipped-label, and diffusion adversarial examples}

Perturbation-based adversarial attacks typically generate adversarial examples by iteratively adding adversarial gradients to clean images, which inevitably introduces noisy patterns. These patterns create visible defects that can be detected by humans. However, these perturbations are applied at the pixel level, leaving the content of the clean image unchanged. In contrast, GAN-based unrestricted adversarial attacks create Unrestricted Adversarial Examples (UAEs) by perturbing the latents. The generator then produces images based on these GAN latents. This method introduces perturbations at the content level, as GAN-based techniques do not directly add noise to the final images. Given the generator's sensitivity to changes in low-dimensional latents, adversarial latents can result in images with entirely different content. This can even lead to a change in the label of the adversarial images, creating what we refer to as flipped-label images.

Adversarial examples generated by diffusion models follow a diffusion generation process, which can be seen as a denoising process. As a result, the noisy gradients injected are removed during the generation process. This necessitates a larger Projected Gradient Descent (PGD) gradient in previous works to successfully generate a UAE, often resulting in a decrease in image quality. In our work, we inject the adversarial objective in an interpretable manner by increasing the conditional likelihood on the target attack label, following the diffusion process. We provide detailed proof of the effectiveness of our adversarial guidance in Appendix A and B. Consequently, our proposed AdvDiff method is more reliable in generating high-quality adversarial examples than simply conducting the PGD attack on the sampled images of the diffusion model.

\section{Ethics Concerns}

AdvDiff can bring security problems to existing DL-based applications, and it generates visually indistinguishable adversarial examples to humans while deceiving the target DL model. This characteristic makes the AdvDiff's images hard to detect by current defense mechanisms, even with human experts. However, our unrestricted adversarial examples can be adopted for adversarial training because our adversarial examples are generated close to the decision boundary of the target classifier. Another critical reason for achieving adversarial training is that the generated adversarial examples have high fidelity and high diversity on the large-scale dataset. Therefore, AdvDiff can have positive social impacts on improving the AI model robustness. 

\section{Limitations}

Although AdvDiff shows superior performance on the unrestricted adversarial attack with large-scale datasets, the generation speed of adversarial examples with diffusion models is relatively slower than GAN-based models. This limitation makes AdvDiff hard to perform a real-time attack. However, the unrestricted adversarial attack does not have a real-time attack scenario. And we can also adopt a fast-sampling method to improve the sampling speed of the AdvDiff, which we aim to improve in future work. Another limitation is that AdvDiff is sensitive to the parameter settings of two adversarial guidance scales $a$ and $s$. The reason is that AdvDiff can deploy in any conditional diffusion model, which has different sampling mechanisms in other datasets. Therefore, we should set the adversarial guidance scales accordingly, but the attack performances are not vastly changed if the scales are in an appropriate range.

\bibliographystyle{splncs04}
\bibliography{main}